# Demystifying Ten Big Ideas and Rules Every Fire Scientist & Engineer Should Know About Blackbox, Whitebox & Causal Artificial Intelligence

M.Z. Naser, PhD, PE
School of Civil and Environmental Engineering, and Earth Sciences, Clemson University, Clemson, SC 29634, USA
AI Research Institute for Science and Engineering (AIRISE), Clemson University, Clemson, SC 29634, USA
E-mail: mznaser@clemson.edu, Website: www.mznaser.com

*Synopsis*
Artificial intelligence (AI) is paving the way towards the fourth industrial revolution with the fire domain (Fire 4.0). As a matter of fact, the next few years will be elemental to how this technology will shape our academia, practice, and entrepreneurship. Despite the growing interest between fire research groups, AI remains absent of our curriculum, and we continue to lack a methodical framework to adopt, apply and create AI solutions suitable for our problems. The above is also true for parallel engineering domains (i.e., civil/mechanical engineering), and in order to negate the notion of "history repeats itself" (e.g., look at the continued debate with regard to modernizing standardized fire testing, etc.), it is the motivation behind this letter to the Editor[1] to demystify some of the big ideas behind AI to jump-start prolific and strategic discussions on the front of *AI & Fire*. In addition, this letter intends to explain some of the most fundamental concepts and clear common misconceptions specific to the adoption of AI in fire engineering. This short letter is a companion to the *Smart Systems in Fire Engineering* special issue sponsored by *Fire Technology*. An in-depth review of AI algorithms [1] and success stories to the proper implementations of such algorithms can be found in the aforenoted special issue and collection of papers.

This letter comprises two sections. The first section outlines *big ideas* pertaining to AI, and answers some of the burning questions with regard to the merit of adopting AI in our domain. The second section presents a set of *rules* or technical *recommendations* an AI user may deem helpful to practice whenever AI is used as an investigation methodology. The presented set of rules are complementary to the big ideas.

*Big ideas.*
The listed big ideas herein are tailored to present key concepts behind AI as a domain and method of investigation.

*Big Idea 1: What is AI?*
A good start to this letter is by defining AI (as well as its derivatives, machine learning (ML), and deep learning (DL)). From this view, The Oxford Dictionary defines AI as *"the theory and development of computer systems able to perform tasks normally requiring human intelligence, such as visual perception, speech recognition, decision-making, and translation between languages"* [2]. In fire engineering terms, AI is the development of fundamental principles that enable computing machines from carrying out tasks that require engineering intellect (i.e., identify failure modes in fire-exposed beams as shear-dominant or flexure-dominant). The same dictionary also defines ML as "*a type of artificial intelligence in which computers use huge amounts of data to learn how to do tasks rather than being programmed to do them*" [2]. In a way, ML encompasses the development of codes/algorithms that enable machines to learn directly from data (e.g., algorithm X, once applied to a set of data collected from combustion experiments, can help build a surrogate model capable of identifying flammability limits of materials). Similarly, deep learning

---

[1] This letter is inspired in part by TZ Harmathy's work "*Ten rules of fire endurance rating*" [31] (also published at Fire Technology).

(DL) is a technique that can be applied to enable ML. DL differs than other algorithms since it learns to process data through a series of processing layers designed to mimic how the brain processes data [3]. Figure 1 showcases the realm of AI, ML, and DL. This figure is bound to change in the coming years as ML and DL flourish into their own disciplines.

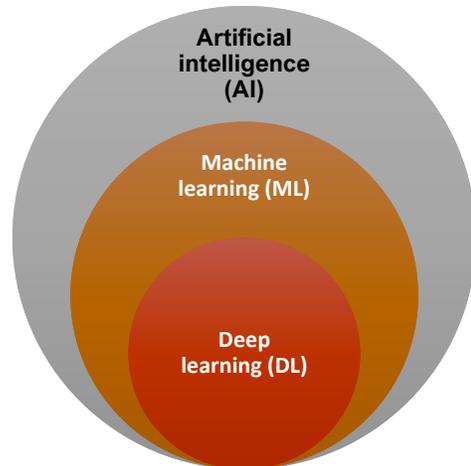

*Big idea 2: How does AI/ML/DL differ from data science and statistics?*
Generally speaking, data science comprises scientific methods and knowledge domain to extract knowledge from data. This data can be clean or noisy, structured or unstructured, real or synthetic. Statistics is a sub-area of data science. Statistical methods infer from a sample of a population with a "relatively" small number of parameters. In a statistical analysis, a model often accompanies a pre-determined statistical distribution to fit the input parameters to the result of a phenomenon [4]. Once the number of parameters grows, or the relationship between them turns complicated, or the data does not satisfy assumptions/conditions tied with statistical methods, such methods become less effective [5]. Unlike statistical methods, AI derivatives make minimal assumptions about data type and directly learn from data in search of patterns (thereby becoming useful on unstandardized data or that containing highly nonlinear interactions). A unique distinction between statistical methods and AI derivatives is that these derivatives need to satisfy a penalization function to attain convergence, as opposed to a set of rules [6].

Fig. 1 A look into AI, ML and DL

*Big idea 3: Why do we need AI/ML/DL in our domain? And why do we need them now?*
The truth of the matter is that most data in our field, and just like other fields, is highly nonlinear, highly dimensional, contains noise and outliners. Thus, much of the real-world data may not easily be processed through traditional methods, and hence engineers tend to smoothen such data via assumptions, etc. Furthermore, classical methods (i.e., human-powered or numerical modeling) primarily require specialized software and computing stations. On the other hand, AI derivatives can be integrated with open-source and free-to-use software. Such derivatives provide orders of magnitude faster solutions and thus can provide us with real-time means to process data and make decisions. One can think of AI derivatives as new methods of investigation that can help *supplement* those of traditional nature – as supposed to <u>replace</u> them. The addition of such tools will expand our arsenal.

*Big Idea 4: How does AI differ from testing or numerical simulations?*
We design tests to examine phenomena [7]. Naturally, tests (or experiments) are limited in size and scope to comply with available testing facilities. The primary goal of our experiments is to tie a cause(s) to an effect(s); or, holistically speaking input(s) to output(s). To maintain control, one parameter is often varied at a time to observe its sole influence [8]. In lieu of tests, we also develop numerical models to provide us with affordable means to simulate phenomena [9]. A common practice of developing a numerical model is to first validate its predictions against tests [10, 11]. Even though we lack a standardized procedure to develop or validate numerical models, modeling remains a cornerstone in our domain [12].

Ultimately, an analysis is carried out to reach an *understanding* of the cause-and-effect governing the phenomenon on hand [13]. This *understanding* is often expressed via a formula (or design aid). A formula comes in handy as it: 1) can be easily applied, 2) expresses the relations between the inputs, as well as these parameters and the outcome/response observed, and most importantly, 3) articulates the space of influencing factors upon the phenomenon on hand [14, 15].

Unlike the above methods, the integration of AI can be broadly summed by a three-step linear procedure: 1) collect data on a phenomenon, 2) apply algorithm X, and 3) declare that algorithm X can predict the phenomenon on hand via fitness tests. In many instances, this algorithm is a Blackbox – we do not know why it predicts the way it does, nor how it ties the parameters to the phenomenon on hand! However, we do know that the data primarily drive such an algorithm. A question often arises, how can we ensure that this data-driven approach matches that from physics?

Herein where a *philosophical* look into AI is warranted. Herein also where recent advancements in explainability and interpretability continue to rise – and should be of interest to us [16–18]. At this point, an AI user can take due diligence to verify their AI's model predictions (i.e., via a series of performance metrics, expert judgments, apply explainability and causal inference measures etc. – which is what we already do when we derive equations from tests or validate numerical models), or may opt not to use AI to investigate a particular phenomenon (i.e., say one that is complex or with minimal insights/tests, etc.).

A decision is to be made, can this problem be solved in a data-driven approach? Do we need to solve it via such an approach? And if so, can we accept the limitations of such an approach? In a way, we can think of AI as a tool to help us and, just like other methods of investigation, AI also has its limitations; for now.

*Big idea 5: Which algorithm(s) to use?*
Answering this question is not simple; however, it could perhaps be articulated by the following. AI derivatives can be generally classified into three categories; *supervised*, *unsupervised* and *semi-supervised* learning [19]. The first type of learning is potentially applicable to the majority of problems in fire engineering. In supervised learning, both the inputs and output(s) are known, but the relation between inputs and output(s) is not. Thus, the objective of supervised learning is to realize an accurate *mapping* function capable of predicting the output(s) from the inputs. Supervised learning can be further broken down into regression (when the output variable is numeric, i.e., fire resistance time) and detection/classification (when the output variable is a category, i.e., fuel-controlled/ventilation-controlled fires) problems.

On the other hand, unsupervised learning can come in handy when only the inputs are known without any corresponding output(s). In this type of learning, an algorithm is tasked with discovering the underlying nature of the phenomenon. Unsupervised learning can be grouped into clustering (grouping by behavior, i.e., concrete of dense microstructure is vulnerable to spalling) and association (discover rules that describe large portions of the observations, i.e., columns made of dense concrete have high capacity at ambient conditions and could spall under fire). On the other hand, semi-supervised learning is suited where a large amount of inputs and only some of the output data is available (i.e., can be applied in automatic tagging of images where the user tags some images, say in unlabeled images where flying embers are to be detected).

Going back to our question, once you identify a problem, then you can very likely to land a learning process to examine such a problem. From there, arriving at a series of suitable algorithms turns into an easy exercise (see [1, 20]).

*Big idea 6: Do we have enough data to apply AI/ML/DL?*
While it is true that data in any domain can be limited in size, quality, or representation [21], however, the above overlooks the fact that while we do have data from years of fire research and practice, the data we have access to and can use to train AI/ML/DL models is limited. This creates an opportunity to share our data; noting how sharing data and AI codes/models can be carried out via online repositories such as Mendeley. We need to work together and harmony to build centralized or decentralized databases that are publicly shareable and free for use. A collective effort of stakeholders in our sector can prove effective in facilitating the integration of AI derivatives. Ensuring accessibility and inclusion[2] will allow us to further AI-based solutions to our problems.

*Big Idea 7: How many data points do we normally need to run an AI analysis?*
A historical rule of thumb implies that a minimum dataset can be ten observations per predictor [22]. This has then been appended to 23 observations per predictor [23]. As of this point in time, investigations continue to be carried out to better answer this question. Both data quality, quantity, and range are key considerations for a proper dataset. More data may not always correspond to better models; more "good" data will likely get us there efficiently.

*Big Idea 8: What features to include in a model?*
Features refer to the inputs used in an AI model. Such features could be the width of an element, material type, chemical composition, etc. At a minimum, a fire scientist or engineer should re-visit the open literature to identify the prime features (parameters) governing a phenomenon. For example, one may examine established publications/equations to identify such features. If such features are unknown, then experts can be consulted, or the user may opt to adopt a data-driven approach and examine the importance of each available feature via weightage of algorithmic analysis. A user may also create *fictitious* features via feature handling and feature engineering tools. Careful should be taken when treading the latter.

*Big idea 9: Do fire scientists/engineers need to learn to code?*
Thanks to our colleagues in computer sciences, a multitude of algorithms are not only available for immediate integration but these algorithms have been extensively examined and validated. Such algorithms are also available for free use as they were developed via an open-source "community effort" wherein interested public members worked in harmony to improve the state of such algorithms. Perhaps we can follow a similar trend and develop fire-based algorithms that best suit our domain in the near future. Hence, efforts to modernize our traditional fire engineering curriculum will prove meritorious.

*Big idea 10: A properly validated AI model is not always a casual AI model.*
Say a model achieves impressive scores during training/validation and testing. Say the same model also performs well on established databases [24]. Now also say that when diagnostic, the model shows similar behavior to that expected from physical observations. Yet, this model may still not be one that can articulate the cause-effect relations between the inputs and the output(s). A causal model is one that is created following causality principles. A user is to decide what they seek – a data-driven model or a causal model.

---

[2] This is an important point – especially since many schools with fire engineering courses may not have experimental facilities. Sharing our data and codes will allow less representative schools to be active members of our domain – simply since AI investigation can be carried out by freely available codes (yet, our data may not be freely available) – see *Big Idea 9*.

Figure 2 demonstrates a typical application of an AI-based model where a database is developed from the space of known features. In many instances, a database does not contain all known features nor include those from the space of possible features due to any of the aforenoted limitations. Thus, an AI-based model developed on such features is likely to be a data-driven model. Simply, a model capable of predicting the phenomenon on hand by comparing how the collected features relate to the response (outcome of the phenomenon). Depending on the used features, such models can have a variety of degrees of usefulness. However, such models may also miss the true essence of the problem. This is often mistaken for the cause-&-effect. Supplementing the above model with sensitivity, explainability, interpretability methods is a good practice – since these methods provide us with a look into how a *model* behaves (i.e., effectively converting a blackbox to a whitebox/glass model). However, this should not be mistaken by how the inputs *cause* the phenomenon.

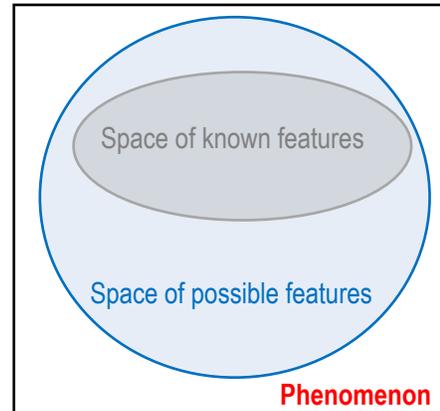

Fig. 2 Data-driven vs. causal models

On the other hand, a causal model can identify the true cause-effect relationships and hence is capable of truly capturing the *essence* of a phenomenon. Such a model goes beyond the space of known features into the space of possible features to articulate how the features interact with each other to yield the phenomenon on hand. The same model can be used to *understand* a phenomenon and explore how to tackle it. The same model also allows us to discover new knowledge[3].

Based on the above Big Ideas, a few *Rules* can be inferred. These rules (or simply *suggestions*) are articulated to allow maximizing the outcome of an AI investigation. These rules revolve around the creation of AI-based models. The specifics to developing AI-based models (in terms of data collection, handling, selection, etc.) are also important, and more on these can be found in traditional textbooks [32, 33] as well as publications [27]. For brevity, these aspects were not touched upon herein, given their generalist nature of being applicable to most, if not all, AI models. These rules were collected based on the recommendation of published works [25–27], observations of current works on the front of fire, and personal experiences.

*Rule 1: Start with simple AI models.*
There is genius in simplicity. Simple AI models are versatile, easy to use, tune and explain, and of low computation needs. As such, it is advisable to start with a simple model vs. complex (advanced) models or ensembles. There is a time and place for complex models to be nurtured or for AI algorithms to be blended into ensembles.

---

[3] For fairness, new knowledge could also be discovered via data-driven models (but not to the same extent as that in causalmodels).

*Rule 2: Explore a series of algorithms, architectures, training/validation methods, and performance metrics.*

As noted earlier, there exist many types of learnings and associated algorithms. Thus, it is of merit to explore how various algorithms will fair against a phenomenon (or more explicitly, will fair against the data on hand) before a model is finalized. In a way, no one algorithm is best suited for all problems. The same can be extended to algorithmic architecture and performance metrics. Correlation does not imply causation. Avoid over tuning models, and watch out for the fine line of tuning for model development vs. model deployment.

*Rule 3: Favor experimental data over synthetic or augmented data alone.*

Real data (i.e., obtained from experiments) may contain randomness effects, which can be attributed to outliers/noise – while in fact, remain part of the physical realm[4] (e.g., sub-phenomena) that we/our equipment cannot fully measure or explain. Yet, such measurements can be valuable (unless proven otherwise). Unlike real data, data obtained by synthetic means (drawn from space of observations) or augmented (i.e., via numerical models) may not capture the aforenoted effects simply due to the idealization associated with such data. For example, running a numerical model with a fixed material model to describe property degradation under fire conditions will inherently cause model predictions (both numerical and AI) to be skewed towards that particular material model [28]. In a way, relying on such data may effectively limit the generalizability of the AI model. The above may, or may not, be of concern (please re-visit *Big Idea 10*).

Perhaps modelers can use a variety of material models (to describe property degradation or mix real with synthetic/augmented data, which may minimize the noted skewness [29]. However, one must remember that replicating randomness effects may not be as easily obtained – especially when high nonlinearity or instability exists. Thus, it is of equal merit to validate and test your model against observations taken from various groups (i.e., testing facilities, equipment, etc.), and across different time spans.

*Rule 4: Explore the inclusion of physics principles into AI models.*

Infusing prior knowledge of physical principles governing our systems is advantageous as such principles regularize the space of solutions and improve the robustness of the transformation functions (often hidden within algorithmic topologies). In addition, the infusing of such principles is a step forward over purely relying on data alone to guide the prediction of an AI model (especially when data is limited, expensive to compile, or incomplete).

*Rule 5: Steer away from Blackbox models (unless the goal is to create a Blackbox model).*

In traditional methods of investigations, we create solutions that help us "understand" the phenomenon on hand. The same is also true whenever AI is used. While Blackbox models have their place, we do need transparent/white/glass AI models that allow us to see and justify model predictions – and most importantly discover new knowledge. A prime goal is to realize *causal* models that can articulate the "cause" and "effect" governing a phenomenon from physics, empirical, and data perspectives [13, 30]. These models will allow a smoother transition toward trusting and embracing AI by our community.

In a way, an equation, expression, or a numerical model can also be thought of as a blackbox. For example, inputting erroneous values into an equation, expression, or a numerical model is expected to yield wrong predictions. Yet, because we can trace how did we arrive at an answer (say, by solving the equation/expression/ matrices/partial differential equations), then we are much more comfortable with accepting such model predictions. The same may not as easily as one can imagine with AI models.

---

4 One can argue that real data from tests may not be representative with real work observations (given the limitation of testing set-ups). This is a valid argument. The second paragraph of *Rule 3* may counter such argument.

*Rule 6: Be cognizant in navigating the trade-offs.*
Similar to other investigation methods, a user may contemplate a few trade-offs[5] while developing an AI model. Some of these trade-offs include those at the fronts of: *bias-variance*, *accuracy-generalizability*, *accuracy-complexity*, *accuracy-computational resources*, *complexity-explainability*, *predictivity-model size*, to name a few (see Fig. 3). While for some, if not most, the adversity of such trade-offs can be minute to our problems, in others, say turbulence analysis, these trade-offs may hinder the development of a model. At the end of the day, achieving a balance between the trade-offs is recommended.

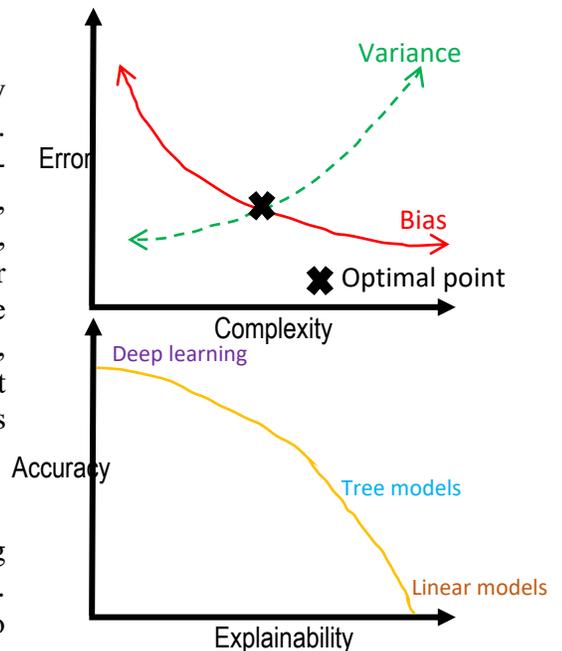

Fig. 3 Examples of AI trade-offs

*Rule 7: Supplement your model with confidence measures.*
Once an AI model completes its training/validation/testing process, such a model is often deployed to the real world. Certainly, real-world data can be different from that used to create this model (in terms of range, quality, etc.). A prime example would be data from wildfires which can be sensitive to geo-spatial or seasonal (or both) effects. Hence, it is of importance that your AI model is able to flag predictions that the model struggles to predict or those likely to be faulty.

*Rule 8: Share your data and code (whenever possible) and full details of your model (at all times).*
This rule builds on our discussion on Big Ideas no. 5 and 6. The same will also facilitate reproducibility and help improve our education on the front of *AI & Fire*. Aspire to create open-source models.

*Rule 9: Transform your AI model into a graphical interface.*
To maximize the outreach of your AI model, such a model can be converted from lines of codes into an App or software with an easy-to-use graphical interface. This may attract users to adopt your model/software – especially that many of our colleagues may not be well versed with AI. At the same time, such software can be copy-righted and marketed for subscriptions. This may indeed bring a new revenue stream to our industry. Finally, such software can also be used in an educational setting to introduce students to AI.

*Rule 10: Keep an eye on your AI model beyond deployment.*
Seek feedback and monitor the performance of the deployed models. Aim to improve and update your model on a regular basis. Use feedback and newly acquired data to enhance your model.

The above *Big Ideas*, and *Rules* are expected to evolve over the next few years rapidly. I hope my colleagues invest in extending and improving these ideas and rules.

---

[5] One can think of a numerical model where dense mesh is used vs. low quality mesh. A trade-off on the front of accuracy-commutation resources is expected to exist.

**Acknowledgment**

I would like to thank the Editor, Prof. Guillermo Rein, and Anonymous reviewer for their support of this work and constructive comments that enhanced the quality of this letter.

**Data Availability**

Some or all data, models, or code that support the findings of this study are available from the corresponding author upon reasonable request.

**Conflict of Interest**

The author declares no conflict of interest.

**References**

1.    Naser MZ (2021) Mechanistically Informed Machine Learning and Artificial Intelligence in Fire Engineering and Sciences. Fire Technol 1–44. https://doi.org/10.1007/s10694-020-01069-8

2.    Oxford English Dictionary (2017) Oxford English Dictionary Online. In: Oxford English Dict.

3.    Lecun Y, Bengio Y, Hinton G (2015) Deep learning. Nature

4.    Christodoulou E, Ma J, Collins GS, et al (2019) A systematic review shows no performance benefit of machine learning over logistic regression for clinical prediction models. J. Clin. Epidemiol.

5.    Bzdok D, Altman N, Krzywinski M (2018) Statistics versus machine learning. Nat Methods. https://doi.org/10.1038/nmeth.4642

6.    Ivezic Ž, Connolly AJ, VanderPlas JT, Gray A (2014) Statistics, Data Mining, and Machine Learning in Astronomy

7.    Christensen BT, Schunn CD (2007) The relationship of analogical distance to analogical function and preinventive structure: The case of engineering design. Mem Cogn. https://doi.org/10.3758/BF03195939

8.    Biot MA (1943) Analytical And Experimental Methods in Engineering Seismology. Trans Am Soc Civ Eng. https://doi.org/10.1061/taceat.0005571

9.    Zienkiewicz OC, Taylor RL (2000) The Finite Element Method Volume 1 : The Basis. Methods

10.    Tekkaya AE (2005) A guide for validation of FE-simulations in bulk metal forming. Arab J Sci Eng

11.    Ferreira J, Gernay T, Franssen J, Vassant O (2018) Discussion on a systematic approach to validation of software for structures in fire - Romeiro Ferreira Joao Daniel. In: SiF 2018: 10th international conference on structures in fire. Belfast

12.    Hawileh RAA, Naser MZZ (2012) Thermal-stress analysis of RC beams reinforced with GFRP bars. Compos Part B Eng 43:2135–2142. https://doi.org/10.1016/j.compositesb.2012.03.004

13.    Pearl J MD (2018) The Book of Why_ The New Science of Cause and Effect-Basic Books

14.    Ellingwood, B., Galambos, T. V., McGregor, J. G. & Cornell CA (1980) Development of a Probability Based Load Criterion for American National Standard A58. U.S. Dep. Commer. Natl. Bur. Stand.

15.    ASCE (2016) Minimum Design Loads for Buildings and Other Structures (ASCE/SEI 7-16)

16.    Rudin C (2019) Stop explaining black box machine learning models for high stakes decisions and use interpretable models instead. Nat. Mach. Intell.

17.    Miller T (2019) Explanation in artificial intelligence: Insights from the social sciences. Artif. Intell.

18.    Naser MZ (2021) An engineer's guide to eXplainable Artificial Intelligence and Interpretable Machine Learning: Navigating causality, forced goodness, and the false perception of inference. Autom Constr 129:103821. https://doi.org/10.1016/J.AUTCON.2021.103821

19.    Bishop C (2006) Pattern recognition and machine learning. Springer

20.    Lattimer BY, Hodges JL, Lattimer AM (2020) Using machine learning in physics-based simulation of fire. Fire Saf J 102991. https://doi.org/10.1016/j.firesaf.2020.102991

21.    Zhu X, Vondrick C, Fowlkes CC, Ramanan D (2016) Do We Need More Training Data? Int J Comput Vis. https://doi.org/10.1007/s11263-015-0812-2

22.    Smeden M van, Moons KG, Groot JA de, et al (2018) Sample size for binary logistic prediction models: Beyond events per variable criteria: https://doi.org/101177/0962280218784726 28:2455–2474. https://doi.org/10.1177/0962280218784726

23.    Riley RD, Snell KIE, Ensor J, et al (2019) Minimum sample size for developing a multivariable prediction model: PART II - binary and time-to-event outcomes. Stat Med. https://doi.org/10.1002/sim.7992

24.    Naser MZ, Kodur V, Thai H-T, et al (2021) StructuresNet and FireNet: Benchmarking databases and machine learning algorithms in structural and fire engineering domains. J Build Eng 102977. https://doi.org/10.1016/J.JOBE.2021.102977

25.    Wujek B, Hall P, Güneş F (2016) Best Practices for Machine Learning Applications. SAS Inst Inc

26.    Artrith N, Butler KT, Coudert FX, et al (2021) Best practices in machine learning for chemistry. Nat. Chem.

27.    Wang AYT, Murdock RJ, Kauwe SK, et al (2020) Machine Learning for Materials Scientists: An Introductory Guide toward Best Practices. Chem Mater. https://doi.org/10.1021/acs.chemmater.0c01907

28.    Naser MZ (2018) Deriving temperature-dependent material models for structural steel through artificial intelligence. Constr Build Mater 191:56–68.

https://doi.org/10.1016/J.CONBUILDMAT.2018.09.186

29. Naser MZ, Kodur VK (2021) Explainable Machine Learning using Real, Synthetic and Augmented Fire Tests to Predict Fire Resistance and Spalling of RC Columns

30. Naser MZ (2021) Mapping functions: A physics-guided, data-driven and algorithm-agnostic machine learning approach to discover causal and descriptive expressions of engineering phenomena. Measurement 185:110098. https://doi.org/10.1016/J.MEASUREMENT.2021.110098

31. Harmathy TZ (1965) Ten rules of fire endurance rating. Fire Technol. https://doi.org/10.1007/BF02588479

32. Barber D (2012) Bayesian reasoning and machine learning

33. Murphy KP (2012) Machine learning: a probabilistic perspective (adaptive computation and machine learning series)